\documentclass[conference]{IEEEtran}
\IEEEoverridecommandlockouts
\usepackage{cite}
\usepackage{amsmath,amssymb,amsfonts}
\usepackage{algorithmic}
\usepackage{graphicx}
\usepackage{textcomp}
\usepackage{xcolor}
\usepackage[caption=false]{subfig}
\usepackage{booktabs}
\usepackage{multirow}
\def\BibTeX{{\rm B\kern-.05em{\sc i\kern-.025em b}\kern-.08em
    T\kern-.1667em\lower.7ex\hbox{E}\kern-.125emX}}
\begin{document}

\title{CORE: Cyclic Orthotope Relation Embedding for Knowledge Graph Completion\\
}

\author{\IEEEauthorblockN{1\textsuperscript{st} Yingqi Zeng}
~\\
\and
\IEEEauthorblockN{2\textsuperscript{nd} Luying Wang}
~\\
\and
\IEEEauthorblockN{3\textsuperscript{rd} Huiling Zhu*}
*Corresponding author

}

\maketitle

\begin{abstract}
Knowledge graph completion (KGC) aims to automatically infer missing facts in multi-relational data by mapping entities and relations into continuous representation spaces. Recent region-based embedding models have shown great promise in capturing complex logical patterns by representing relations as geometric regions. However, these models inevitably suffer from absolute boundary constraints during optimization. Conversely, without such constraints, relation regions expand indefinitely. To address the limitation, we propose \textbf{CORE} (Cyclic Orthotope Relation Embedding), a novel KGC model that embeds entities and relations onto a boundary-less torus manifold.CORE represents relations as cyclic orthotopes on the torus manifold, allowing regions to seamlessly wrap around spatial boundaries to ensure smooth gradient conduction. Furthermore, an adaptive width regularization is introduced to prevent unconditional region expansion. Theoretical analysis proves that CORE can capture various complex relation patterns such as subsumption and intersection. Extensive experiments on four benchmark datasets demonstrate that CORE achieves highly competitive performance, significantly improving link prediction accuracy in dense semantic environments.
\end{abstract}


\section{Introduction}

With the rapid development of information technology, the volume of data generated by human society is growing exponentially. Extracting valuable structural knowledge from massive, heterogeneous, and unstructured data has become a core challenge in the field of artificial intelligence. Knowledge Graphs (KGs) have emerged as an effective solution to this demand. Fundamentally, a knowledge graph is a semantic network that reveals the relations between entities. It employs a graph structure to store real-world facts, where nodes represent real-world entities and edges represent relations between them. The fundamental format of KGs are factual triples, denoted as $(h, r, t)$, indicating that a head entity $h$ is connected to a tail entity $t$ via a relation $r$. Owing to their structured nature, large-scale KGs such as Freebase\cite{Freebase}, DBpedia\cite{DBpedia}, YAGO\cite{YAGO} and Wikidata \cite{Wikidata} have been widely deployed in downstream applications, including intelligent question answering\cite{zhang2021} and recommendation systems\cite{CKBE}. 

Despite their massive scale, knowledge graphs are often incomplete. Since KGs are primarily constructed through automated or semi-automated extraction algorithms from unstructured text or encyclopedias, they suffer from algorithmic inaccuracies and the inherent latency of knowledge. This incompleteness severely restricts the reasoning capabilities and generalization potential of KGs in downstream tasks. Consequently, Knowledge Graph Completion (KGC), also known as link prediction, has become a critical research area. The primary objective of KGC is to automatically infer missing fact such as predicting the tail entity in $(h, r, ?)$ or the relation in $(h, ?, t)$ based on existing triples within the graph. 

To tackle the KGC task, Knowledge Graph Embedding (KGE) has become the mainstream paradigm. Traditional translational distance models, such as TransE \cite{TransE} and RotatE \cite{RotatE}, project entities and relations into continuous vector spaces, defining relations as translations or rotations. To enhance expressive power, region-based embedding models have been proposed. Models like BoxE \cite{BoxE} embed entities as points but model relations as geometric regions. A triple is considered valid if the entities fall within the corresponding relation's spatial region, allowing the model to naturally capture complex mapping properties and hierarchical structures.

However, existing region-based embedding models all face the same challenge. They inevitably suffer from absolute boundary constraints within the representation space. During training and optimization processes, if absolute spatial boundaries are not enforced, relational regions tend to expand indefinitely to accommodate more entities, while negative samples may diverge without boundaries. Conversely, imposing forced spatial mapping or hard boundary truncation to constrain the embedding space may induce severe gradient saturation at the edges of relational regions. This hard truncation approach mathematically distorts the geometric structure of relations and excludes boundary entities that logically belong to them.Consequently, the model's ability to precisely characterize relation regions is compromised, heavily limiting the overall completion performance.

To overcome the boundary limitations of traditional region embeddings, we propose \textbf{CORE} (Cyclic Orthotope Relation Embedding), a novel KGC model based on spatial representation. CORE extends region embeddings into the torus manifold, a compact Abelian Lie group. The model represents entities as dynamic points and relations as boundary-less cyclic orthotopes. Specifically, a factual triple is considered true if both the head and tail entities fall precisely within their corresponding cyclic orthotopes. By leveraging the equivalence class mapping rules of the quotient space, CORE allows relation regions to seamlessly cross spatial boundaries while maintaining continuity. This topological advantage effectively resolves the local gradient saturation at region edges, capturing boundary entities that would otherwise be discarded in traditional models. Furthermore, to prevent the relation regions from falling into an inflation shortcut within the boundary-less manifold, we introduce an adaptive width regularization mechanism to maintain high precision.

The main contributions of this paper are summarized as follows:

\begin{itemize}
	\item We introduce region embeddings into a boundary-less torus manifold to address the boundary constraints prevalent in previous methods. Specifically, we represent entities as dynamic points and relations as cyclic orthotopes. Furthermore, we apply an adaptive width regularization strategy to prevent unconditional region inflation.
	\item Our model captures various complex relation patterns through geometric interactions between cyclic orthotope regions. We theoretically demonstrate the model's capability to express patterns including symmetry, anti-symmetry, inversion, subsumption, intersection, and mutual exclusion.
	\item We evaluate the performance of our model on the link prediction task using the FB15k, FB15k-237, WN18, and WN18RR benchmark datasets. Experimental results show that CORE achieves competitive performance against various baseline models.
\end{itemize}

\section{Related Work}

\subsection{Representation Space-based Models}
Traditional KGC models predominantly learn embeddings by mapping entities and relations into specific mathematical spaces to capture relational patterns and structural attributes.

\textbf{Models in Common Vector Spaces.} The most widely adopted approach involves embedding knowledge graphs into real or complex vector spaces. TransE \cite{TransE} is a pioneering model that represents relations as simple translations between entity vectors. To address its limitations in modeling complex relations, subsequent variants like TransH \cite{TransH} and TransR \cite{TransR} introduced relation-specific hyperplanes and separate relation spaces. Alternatively, semantic matching models such as DistMult \cite{DistMult} utilize bilinear operations to capture semantic interactions. To properly model asymmetric relations, ComplEx \cite{Complex} extends the embeddings into the complex vector space, while RotatE \cite{RotatE} elegantly defines relations as rotations in the complex plane. SimplE \cite{SimplE} models asymmetric relations by learning two independent representations for each entity and two diagonal matrices for each relation. DualE \cite{DualE} embeds entities and relations into the dual quaternion space, a hypercomplex extension of the complex number system. HAKE \cite{HAKE} maps entities
into the polar coordinate system such that it can distinguish the hierarchical
levels of the entities.

\textbf{Models in other Spaces.} Beyond standard vector spaces, researchers have explored specialized geometric and algebraic structures to better reflect the underlying topology of knowledge graphs. For instance, TorusE \cite{TorusE} and KGLG \cite{KGLG} map entities onto a torus (a compact Lie group) to naturally resolve the divergence issues of embeddings. To capture the inherent hierarchical and tree-like structures of KGs, hyperbolic geometry models such as MuRP \cite{MuRP} and ATTH \cite{ATTH} leverage negative curvature spaces. Additionally, spherical geometries, as seen in TransC \cite{TransC}, have been utilized to encapsulate concepts and instances within closed manifolds.

\textbf{Region-based Embedding Models.} As a highly expressive subset of representation space models, region-based embeddings shift the paradigm by modeling entities as points and relations as complex geometric regions. A triple is scored based on the distance from the entity points to the relation regions. BoxE \cite{BoxE}  pioneers this by modeling relations as hyper-rectangles in Euclidean space, giving entities dynamic contextual representations based on their interacting counterparts. To capture more complex compositional rules, ExpressivE \cite{ExpressivE}  expands this concept by defining relations as hyper-parallelograms in a virtual triple space. Similarly, Octagon \cite{Octagon}  improves geometric intuitiveness by utilizing axis-aligned octagons to enforce boundary constraints. While these region-based models suffer from absolute boundary constraints. Our proposed model, CORE, belongs to this category but critically diverges by mapping cyclic orthotope regions onto a boundary-less torus manifold, thereby eliminating the gradient saturation issues associated with absolute boundaries.


\subsection{Neural Network-based Models}
To overcome the limited expressiveness of shallow representation spaces in modeling highly non-linear interactions, neural networks have been widely applied to KGC. ConvE \cite{ConvE} introduces 2D convolutional layers over reshaped embeddings to extract global semantic features efficiently with fewer parameters. Furthermore, Graph Neural Networks (GNNs) have been adopted to aggregate structural neighborhood information. For example, R-GCN \cite{R-GCN} introduces relation-specific transformations for message passing, while KBGAT \cite{KBGAT} incorporates multi-head attention mechanisms to weigh the importance of different neighboring nodes. Although highly expressive, neural network models often face challenges related to high computational complexity, risk of overfitting, and lack of interpretability.

\subsection{Pre-trained Language Model-based Models}
With the rapid advancement of Large Language Models (LLMs), incorporating textual context and external knowledge into KGC has yielded significant performance gains. Models like KG-BERT \cite{KG-BERT} treat triples as text sequences, framing link prediction as a sequence classification task encoded by PLMs. Extensions like SimKGC \cite{SimKGC} introduce contrastive learning to enhance the model's discriminative reasoning abilities. Other approaches, such as KGT5 \cite{KGT5} and KICGPT \cite{KICGPT}, utilize generative architectures and in-context learning to bridge structural knowledge with natural language. While these methods demonstrate exceptional performance by leveraging vast external corpora, their heavy reliance on massive parameters makes them computationally expensive and occasionally susceptible to model hallucinations.

\section{Method}
\label{sec:method}

\subsection{Notations}
Let a knowledge graph be defined as a directed graph $\mathcal{G}=\{\mathcal{E}, \mathcal{R}, \mathcal{T}\}$, where $\mathcal{E}$ is the set of entities, $\mathcal{R}$ is the set of relations, and $\mathcal{T} \subseteq \mathcal{E} \times \mathcal{R} \times \mathcal{E}$ is the set of facts. A factual triple is denoted as $r(e_h, e_t)$, where $e_h, e_t \in \mathcal{E}$ act as the head and tail entities, and $r \in \mathcal{R}$ is the relation between them. 

The target embedding space of CORE is a $d$-dimensional torus $\mathbb{T}^d \cong [0, 1)^d$, with $[0, 1)$ serving as its fundamental domain. We use $\mathbf{e}_h^{r(e_h, e_t)}$ to denote the specific dynamic representation of the head entity in the context of the triple $r(e_h, e_t)$. $\|\cdot\|_x$ denotes the $L_x$-norm, while $\odot$ and $\oslash$ represent element-wise multiplication and division, respectively.

\subsection{Preliminaries: Torus}
To support boundary-less spatial mapping and continuous gradient descent, CORE selects the compact Abelian Lie group as its underlying mathematical space. Specifically, entities and relations are embedded into an $n$-dimensional torus space $\mathbb{T}^n$. Mathematically, the torus $\mathbb{T}^n$ is defined as the quotient space of the real coordinate space $\mathbb{R}^n$ under the integer lattice $\mathbb{Z}^n$:
\begin{equation}
	\mathbb{T}^n \cong \mathbb{R}^n / \sim = \{[\mathbf{x}] \mid \mathbf{x} \in \mathbb{R}^n\},
\end{equation}
where $[\mathbf{x}]$ represents an equivalence class. In practical computations, any diverging coordinate $\mathbf{x} \in \mathbb{R}^n$ can be uniquely mapped to standard coordinates within $[0, 1)^n$ via a non-negative modulo operation ($\mathbf{x} \bmod 1$). This periodic wraparound property ensures that when an entity moves and crosses the spatial bounds, it seamlessly re-enters from the opposite side, providing mathematical support for smooth distance evaluation.

\subsection{The CORE Model}

\subsubsection{Dynamic Entity Representation}
To enable entities to exhibit varying semantic features when interacting with different objects, we introduce a context-aware dynamic representation mechanism. Every entity $e_i \in \mathcal{E}$ is parameterized by two vectors: a base position $\mathbf{e}_i \in \mathbb{R}^d$ representing its initial coordinate on the torus, and a transformational bump $\mathbf{b}_i \in \mathbb{R}^d$ denoting its spatial shifting effect on other entities.

For a given triple $r(e_h, e_t)$, the final embeddings of the head and tail entities are influenced by each other's transformational bumps. Under the natural projection rule of the torus geometry, the final dynamic embeddings are calculated as:
\begin{align}
	\mathbf{e}_h^{r(e_h, e_t)} &= (\mathbf{e}_h + \mathbf{b}_t) \bmod 1, \\
	\mathbf{e}_t^{r(e_h, e_t)} &= (\mathbf{e}_t + \mathbf{b}_h) \bmod 1.
\end{align}
This mechanism allows a single entity to flexibly distribute across different local regions of the torus topology depending on the relational context.

\subsubsection{Cyclic Orthotope Relation Representation}
To complement the dynamic entity embeddings, CORE models each relation $r \in \mathcal{R}$ as two independent cyclic orthotopes—on $\mathbb{T}^d$: the head region $\mathbf{r}^h$ and the tail region $\mathbf{r}^t$. 

Taking the head region $\mathbf{r}^h$ as an example, its spatial span is determined by two core parameters: the region center $\mathbf{c}_{\mathbf{r}^h} \in [0, 1)^d$, which acts as the semantic anchor, and the region width $\mathbf{w}_{\mathbf{r}^h} \in (0, 0.5]^d$, which dictates the coverage capacity along each dimension. The theoretical upper limit of the width is naturally constrained to $0.5$ because the maximum topological distance in $\mathbb{T}^d$ is $0.5$.

Combining these parameters with the torus distance metric, the set of points $B_{\mathbf{r}^h}$ covered by the cyclic orthotope is formulated as:
\begin{equation}
	B_{\mathbf{r}^h} = \left\lbrace \mathbf{x} \in \mathbb{T}^d \mid \min(|\mathbf{x} - \mathbf{c}_{\mathbf{r}^h}|, \mathbf{1} - |\mathbf{x} - \mathbf{c}_{\mathbf{r}^h}|) \le \mathbf{w}_{\mathbf{r}^h} \right\rbrace.
\end{equation}

\begin{figure}[htbp]
	\centering
	\subfloat[2D Unrolled Plane]{%
		\label{fig:CORE_a}
		\includegraphics[width=0.6\linewidth]{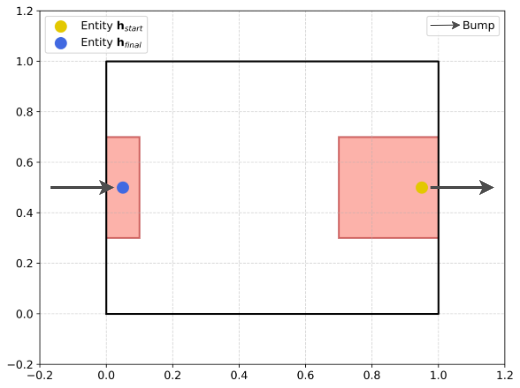}
	}
	\vspace{2em} 
	
	\subfloat[3D Torus Manifold Perspective]{%
		\label{fig:CORE_b}
		\includegraphics[width=0.6\linewidth]{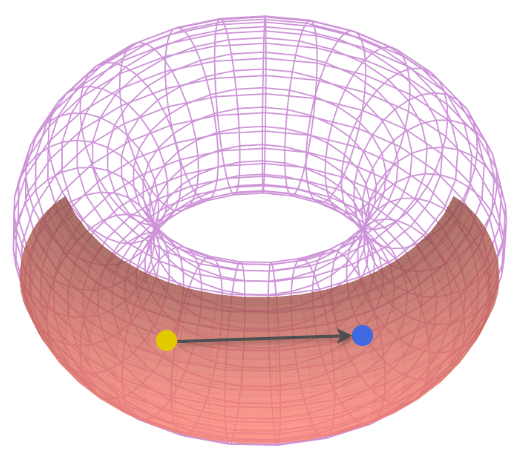}
	}
	\caption{Illustration of the CORE model. The cyclic orthotope naturally wraps around the spatial boundaries.}
	\label{fig:CORE-model}
\end{figure}

Geometrically, a factual triple $r(e_h, e_t)$ is considered valid if the final transformed embedding of the head entity $\mathbf{e}_h^{r(e_h, e_t)}$ falls precisely within the boundary of the relation's head region $\mathbf{r}^h$, with a similar condition applying to the tail entity. The most significant advantage of modeling relation regions as cyclic orthotopes lies in their natural topological wraparound property. To clearly illustrate the topological interaction between entities and cyclic orthotopes, Figure \ref{fig:CORE-model} provides an intuitive example from both a 2D unrolled plane and a 3D manifold perspective.

Consider a valid triple that crosses the spatial boundary, where the final embedding of the head entity is denoted as $\mathbf{h}_{final}$. In traditional region-based models, when the region width expands and touches the artificially imposed boundaries (e.g., the right edge in Figure \ref{fig:CORE_a}) due to spatial restrictions, the exceeding portion is forcefully truncated. This hard truncation severely impairs the model's expressive capacity and logical integrity. 

In contrast, within the CORE space, when the upper bound of the head region $\mathbf{r}^h$ exceeds the coordinate $1.0$, the exceeding portion strictly follows the equivalence class mapping rule and seamlessly re-enters the space from the lower bound $0.0$ on the left side. Consequently, the entity $\mathbf{h}_{final}$, which lands on the extreme left of the space, is perfectly enveloped within the logically continuous head region $\mathbf{r}^h$ under the torus topology—despite being numerically distant from the region's center on the right. The model successfully and naturally evaluates this as a valid fact.

As visualized in the 3D torus manifold in Figure ~\ref{fig:CORE_b}, this boundary-crossing region is, in its physical essence, an absolutely continuous surface patch. This seamless nature of the relation region plays a pivotal role in model optimization: it ensures that for any entity located within a cross-boundary region, the derivatives remain absolutely continuous and smooth as it is optimized toward the region center, effectively eliminating the gradient saturation and vanishing issues prevalent at the edges of Euclidean spaces.

\subsubsection{Distance Metric and Scoring Function}
For any two points $\mathbf{x}, \mathbf{y}$ in the torus space, the shortest topological distance on the $i$-th dimension is defined as $\delta(x_i, y_i) = \min(|x_i - y_i|, 1 - |x_i - y_i|)$. 

To quantify the spatial plausibility of an entity embedding $\mathbf{x}$ falling into a target relation region $\boldsymbol{r}$, we design a smooth, piecewise distance function:
\begin{equation}
	\mathrm{dist}(\mathbf{x}, \boldsymbol{r}) =
	\begin{cases}
		\delta(\mathbf{x}, \mathbf{c}_{\boldsymbol{r}}) \oslash \mathbf{w}_{\boldsymbol{r}}, & \text{if } \delta(\mathbf{x}, \mathbf{c}_{\boldsymbol{r}}) \le \mathbf{w}_{\boldsymbol{r}}, \\[1.5ex]
		(\delta(\mathbf{x}, \mathbf{c}_{\boldsymbol{r}}) - \mathbf{w}_{\boldsymbol{r}}) \oslash \mathbf{w}_{\boldsymbol{r}}^2 + \mathbf{1}, & \text{otherwise.}
	\end{cases}
\end{equation}
This function provides gentle gradients to allow entities to traverse freely within the region, while applying sharp, quadratic penalties to negative samples outside the region. 

The overall distance $d(r(e_h, e_t))$ for a triple is the aggregated spatial distance of both the head and tail entities (utilizing norms such as $L_1$, $L_2$, or $eL_2$). Finally, the scoring function is defined as $f(r(e_h, e_t)) = -d(r(e_h, e_t))$, where a higher score implies a higher probability of the triple being factual.

\subsection{Model Optimization and Width Regularization}
To train the model effectively, we employ a margin-based self-adversarial negative sampling loss function \cite{RotatE}. This loss function assigns higher exponential weights to difficult negative samples (i.e., incorrect triples located close to the region centers). 

Furthermore, since the fundamental domain of $\mathbb{T}^d$ is bounded within $[0, 1)^d$, standard region-based training often exploits an optimization shortcut: unconditionally inflating the relation widths $\mathbf{w}_{\mathbf{r}^h}$ and $\mathbf{w}_{\mathbf{r}^t}$ to cover the entire manifold. To suppress this risk, we introduce an adaptive width regularization penalty using the $L_2$-norm:
\begin{equation}
	L_{Reg} = \frac{1}{|\mathcal{R}|} \sum_{r \in \mathcal{R}} \left( \|\mathbf{w}_{\mathbf{r}^h}\|_2^2 + \|\mathbf{w}_{\mathbf{r}^t}\|_2^2 \right).
\end{equation}
The final joint optimization objective is formulated as $L = L_{KGE} + \lambda L_{Reg}$, where $\lambda \ge 0$ is a hyperparameter balancing the regularization strength.

\subsection{Model Capabilities}
Based on geometric interactions on the torus space, CORE is theoretically fully expressive and capable of capturing diverse and complex relational patterns:
\begin{itemize}
	\item \textbf{Symmetry:} $r(x,y) \Rightarrow r(y,x)$. CORE captures symmetry by learning identical parameters for the head and tail regions: $\mathbf{c}_{\mathbf{r}^h} = \mathbf{c}_{\mathbf{r}^t}$ and $\mathbf{w}_{\mathbf{r}^h} = \mathbf{w}_{\mathbf{r}^t}$.
	\item \textbf{Anti-symmetry:} $r(x,y) \Rightarrow \neg r(y,x)$. CORE models this by ensuring the head and tail regions are strictly disjoint in the topological space: $\mathbf{r}^h \cap \mathbf{r}^t = \emptyset$.
	\item \textbf{Inversion:} $r_1(x,y) \iff r_2(y,x)$. This is achieved by cross-sharing region parameters between the two relations: $(\mathbf{r}_1^h = \mathbf{r}_2^t) \land (\mathbf{r}_1^t = \mathbf{r}_2^h)$.
	\item \textbf{Subsumption:} $r_1(x,y) \Rightarrow r_2(x,y)$. Modeled by completely enveloping the child relation's region within the parent relation's region: $(\mathbf{r}_1^h \subseteq \mathbf{r}_2^h) \land (\mathbf{r}_1^t \subseteq \mathbf{r}_2^t)$.
	\item \textbf{Intersection:} $r_1(x,y) \land r_2(x,y) \Rightarrow r_3(x,y)$. CORE captures logical derivation by ensuring the geometric overlap of prior relations is contained within the target relation: $((\mathbf{r}_1^h \cap \mathbf{r}_2^h) \subseteq \mathbf{r}_3^h) \land ((\mathbf{r}_1^t \cap \mathbf{r}_2^t) \subseteq \mathbf{r}_3^t)$.
	\item \textbf{Mutual Exclusion:} $r_1(x,y) \land r_2(x,y) \Rightarrow \perp$. Modeled by guaranteeing that either the head or tail regions of the two relations have zero overlap: $(\mathbf{r}_1^h \cap \mathbf{r}_2^h = \emptyset) \lor (\mathbf{r}_1^t \cap \mathbf{r}_2^t = \emptyset)$.
\end{itemize}

\section{Experiments}
\label{sec:experiments}

To evaluate the performance of the proposed CORE model on the Knowledge Graph Completion (KGC) task, we conduct extensive link prediction experiments on multiple benchmark datasets.

\subsection{Experimental Setup}

\subsubsection{Datasets}
We evaluate our model on four widely used public KGC benchmarks: FB15k \cite{TransE}, FB15k-237 \cite{FB15K-237}, WN18 \cite{TransE}, and WN18RR \cite{ConvE}. The detailed statistics of these datasets are summarized in Table \ref{tab:datasets}.

\begin{table}[htbp]
	\centering
	\caption{Statistical details of the benchmark datasets}
	\label{tab:datasets}
	\begin{tabular}{lccccc}
		\toprule
		\textbf{Dataset} & \textbf{\# Entities} & \textbf{\# Relations} & \textbf{\# Train} & \textbf{\# Valid} & \textbf{\# Test} \\
		\midrule
		FB15k       & 14,951  & 1,345 & 483,142   & 50,000 & 59,071 \\
		FB15k-237   & 14,541  & 237   & 272,115   & 17,535 & 20,466 \\
		WN18        & 40,943  & 18    & 141,442   & 5,000  & 5,000  \\
		WN18RR      & 40,943  & 11    & 86,835    & 3,034  & 3,134  \\
		\bottomrule
	\end{tabular}
\end{table}

\subsubsection{Evaluation Metrics}
The core objective of link prediction is to infer the missing head entity $(?, r, t)$ or tail entity $(h, r, ?)$. Following standard evaluation protocols in this domain, we report results under the "Filtered" setting, meaning that all genuine triples present in the training, validation, and test sets are removed from the candidate list prior to ranking to avoid penalizing correct predictions. We evaluate the predictive performance using Mean Reciprocal Rank (MRR) and Hits@$K$ ($K \in \{1, 3, 10\}$).

MRR computes the average of the reciprocal ranks of the correct entities. A higher MRR indicates that the model ranks the correct entities closer to the top. Let $|\mathcal{S}|$ be the total number of test queries and $\mathrm{rank}_i$ be the rank of the correct target entity for the $i$-th query:
\begin{equation}
	\mathrm{MRR} = \frac{1}{|\mathcal{S}|} \sum_{i=1}^{|\mathcal{S}|} \frac{1}{\mathrm{rank}_i}.
\end{equation}

Hits@$K$ measures the proportion of correct entities ranked in the top $K$ positions. $\mathbb{I}(\cdot)$ is an indicator function that returns 1 if the condition is met and 0 otherwise:
\begin{equation}
	\mathrm{Hits}@K = \frac{1}{|\mathcal{S}|} \sum_{i=1}^{|\mathcal{S}|} \mathbb{I}(\mathrm{rank}_i \le K).
\end{equation}

\subsubsection{Baselines}
To verify the effectiveness of our approach, we compare CORE against several robust baselines, categorized by their representation mechanisms: 
1) \textit{Translational Distance Models}: TransE \cite{TransE}, TransR \cite{TransR}, RotatE \cite{RotatE}, DualE \cite{DualE}, HAKE \cite{HAKE}. 
2) \textit{Semantic Matching (Tensor) Models}: DistMult \cite{DistMult}, ComplEx \cite{Complex}, SimplE \cite{SimplE}. 
3) \textit{Torus-based Models}: TorusE \cite{TorusE}, KGLG \cite{KGLG}. 
4) \textit{Neural Network Models}: R-GCN \cite{R-GCN}, ConvE \cite{ConvE}. 
5) \textit{Region-based Models}: BoxE \cite{BoxE}, ExpressivE \cite{ExpressivE}, Octagon \cite{Octagon}.

\subsubsection{Hyperparameters}
Based on grid search, the optimal configurations for CORE are determined as follows: we utilize the Adam optimizer, embedding dimension $d = 500$, batch size $= 512$, negative samples per positive sample $= 1024$, self-adversarial sampling temperature $\alpha = 0.5$, and fixed margin $\gamma = 9.0$. The distance function relies on $d_{eL_2}$. For all baseline models, we report the best results obtained from their original papers or reproduce them strictly following their recommended configurations.

\subsection{Main Results}
Table \ref{tab:results_wn} presents the link prediction results on the WN18 and WN18RR datasets, while Table \ref{tab:results_fb} displays the results for FB15k and FB15k-237. The best results are highlighted in bold, and the second-best are underlined.

\begin{table*}[t]
	\centering
	\setlength{\tabcolsep}{6pt} 
	\caption{Link Prediction Results on WN18 and WN18RR Datasets}
	\label{tab:results_wn}
	\begin{tabular*}{\linewidth}{@{\extracolsep{\fill}} l *{8}{c} @{}}
		\toprule[1.5pt]
		\multirow{2}{*}{\textbf{Method}} & \multicolumn{4}{c}{\textbf{WN18}} & \multicolumn{4}{c}{\textbf{WN18RR}} \\
		\cmidrule(lr){2-5} \cmidrule(lr){6-9}
		& MRR & H@1 & H@3 & H@10 & MRR & H@1 & H@3 & H@10 \\
		\midrule[1pt]
		TransE      & .397 & .040 & .745 & .923 & .182 & .027 & .295 & .444 \\
		TransR      & .605 & .335 & .876 & .940 & .412 & .398 & .425 & .451 \\
		DistMult    & .822 & .728 & .914 & .936 & .430 & .390 & .440 & .490 \\
		ComplEx     & .941 & .936 & .945 & .947 & .440 & .410 & .460 & .510 \\
		SimplE      & .938 & .901 & .925 & .941 & .398 & .382 & .402 & .427 \\
		R-GCN       & .814 & .686 & .928 & .955 & .123 & .137 & .180 & .207 \\
		ConvE       & .942 & .935 & .947 & .955 & .460 & .390 & .430 & .480 \\
		RotatE      & .949 & .944 & .952 & .959 & .476 & .428 & .492 & .571 \\
		DualE       & \underline{.951} & \underline{.945} & \textbf{.956} & \underline{.961} & .482 & .440 & .500 & .561 \\
		HAKE        & .945 & .926 & .927 & .950 & .497 & .452 & \underline{.516} & \underline{.582} \\
		TorusE      & .947 & .943 & .950 & .954 & .452 & .422 & .464 & .512 \\
		KGLG        & .947 & .943 & .950 & .954 & .464 & .429 & .480 & .534 \\
		BoxE        & - & - & - & - & .451 & .400 & .472 & .541 \\
		ExpressivE  & - & - & - & - & \textbf{.508} & \underline{.464} & \textbf{.522} & \textbf{.597} \\
		Octagon    & - & - & - & - & .479 & .436 & .492 & .561 \\
		\midrule
		\textbf{CORE (Ours)} & \textbf{.952} & \textbf{.950} & \underline{.953} & \textbf{.967} & \underline{.498} & \textbf{.470} & .502 & .575 \\
		\bottomrule[1.5pt]
	\end{tabular*}
\end{table*}

On the WN18 dataset, which contains rich semantic relations, CORE exhibits outstanding performance. Specifically, CORE outperforms all baselines in MRR (0.952), Hits@1 (0.950), and Hits@10 (0.967), while achieving the second-best result in Hits@3. Compared to BoxE—a Euclidean region-based model utilizing hyper-rectangles, CORE achieves substantial improvements (e.g., +2.7\% in MRR and +4.5\% in Hits@1). This validates that eliminating boundary optimization bottlenecks unlocks the immense expressive potential of region embeddings. Furthermore, compared to TorusE (a point-based model on the torus), CORE's superior performance confirms that modeling relations as parameterized continuous regions (cyclic orthotopes) offers greater semantic capacity.

The WN18RR dataset removes easily inferable inverse relations and presents a highly sparse, deep tree-like hierarchical structure (predominantly hypernym/hyponym relations). This poses a severe challenge to a model's capacity to encapsulate hierarchies. On WN18RR, CORE surpasses all baselines in the Hits@1 metric and achieves the second-best MRR, demonstrating exceptional precision in exact hit predictions. While ExpressivE shows strong results here, CORE achieves competitive accuracy without ExpressivE's highly complex computational overhead, rendering it far more parameter-efficient.

\begin{table*}[t]
	\centering
	\setlength{\tabcolsep}{6pt} 
	\caption{Link Prediction Results on FB15k and FB15k-237 Datasets}
	\label{tab:results_fb}
	\begin{tabular*}{\linewidth}{@{\extracolsep{\fill}} l *{8}{c} @{}}
		\toprule[1.5pt]
		\multirow{2}{*}{\textbf{Method}} & \multicolumn{4}{c}{\textbf{FB15k}} & \multicolumn{4}{c}{\textbf{FB15k-237}} \\
		\cmidrule(lr){2-5} \cmidrule(lr){6-9}
		& MRR & H@1 & H@3 & H@10 & MRR & H@1 & H@3 & H@10 \\
		\midrule[1pt]
		TransE      & .414 & .247 & .534 & .688 & .257 & .174 & .284 & .420 \\
		TransR      & .346 & .218 & .404 & .582 & .265 & .183 & .305 & .425 \\
		DistMult    & .654 & .546 & .733 & .824 & .241 & .155 & .263 & .419 \\
		ComplEx     & .692 & .599 & .759 & .840 & .247 & .158 & .275 & .428 \\
		SimplE      & .727 & .600 & .773 & .838 & .162 & .090 & .170 & .317 \\
		R-GCN       & .651 & .541 & .736 & .825 & .248 & .153 & .258 & .417 \\
		ConvE       & .745 & .670 & .801 & .873 & .316 & .239 & .350 & .491 \\
		RotatE      & \textbf{.797} & \underline{.746} & \underline{.830} & \underline{.884} & .338 & .241 & .375 & .533 \\
		DualE       & \underline{.790} & .734 & .829 & .881 & .330 & .237 & .363 & .518 \\
		HAKE        & .714 & .639 & .768 & .834 & \underline{.346} & \underline{.250} & \underline{.381} & \underline{.542} \\
		TorusE      & .733 & .674 & .771 & .832 & .305 & .217 & .335 & .484 \\
		KGLG        & .751 & .703 & .782 & .835 & .307 & .219 & .337 & .485 \\
		BoxE        & - & - & - & - & .337 & .238 & .374 & .538 \\
		ExpressivE  & - & - & - & - & .333 & .243 & .366 & .512 \\
		Octagon     & - & - & - & - & .332 & .241 & .367 & .517 \\
		\midrule
		\textbf{CORE (Ours)} & \underline{.790} & \textbf{.753} & \textbf{.841} & \textbf{.891} & \textbf{.351} & \textbf{.255} & \textbf{.387} & \textbf{.546}\\
		\bottomrule[1.5pt]
	\end{tabular*}
\end{table*}

On FB15k, characterized by a massive volume of inverse relations and highly dense subgraphs, CORE secures absolute superiority across Hits@1, Hits@3, and Hits@10. In evaluating overall ranking quality, CORE ties for second place in MRR with DualE. Notably, when compared to other region-based paradigms like BoxE, CORE establishes a substantial lead (e.g., +6.5\% in MRR and +9.0\% in Hits@3). This indicates that the cyclic relation regions introduced by CORE successfully liberate entity distribution freedom within crowded, common-sense KGs, yielding significant boosts in hit rates across all error-tolerance thresholds.

For FB15k-237, inverse relations that lead to data leakage are entirely removed. Consequently, models can no longer rely on simple rule memorization and must execute deeper combinatorial and hierarchical reasoning. Under these stringent conditions, CORE outperforms all baselines. Specifically, CORE achieves 0.351 in MRR, 0.255 in Hits@1, 0.387 in Hits@3, and 0.546 in Hits@10. This robust performance demonstrates the structural completeness of the CORE architecture. When direct inverse clues are unavailable, CORE expertly leverages geometric subsumption, intersection, and other topological constraints to accurately isolate correct entities from a vast pool of candidates. 

\subsection{Parameter Sensitivity Analysis}
To deeply investigate the sensitivity and effectiveness of the proposed width regularization mechanism, we train the model using different regularization coefficients $\lambda$ while holding all other hyperparameters constant. 

	\begin{figure}[htbp]
		\centering
		\subfloat[WN18]{%
			\label{fig:lambda_WN18}
			\includegraphics[width=0.48\linewidth]{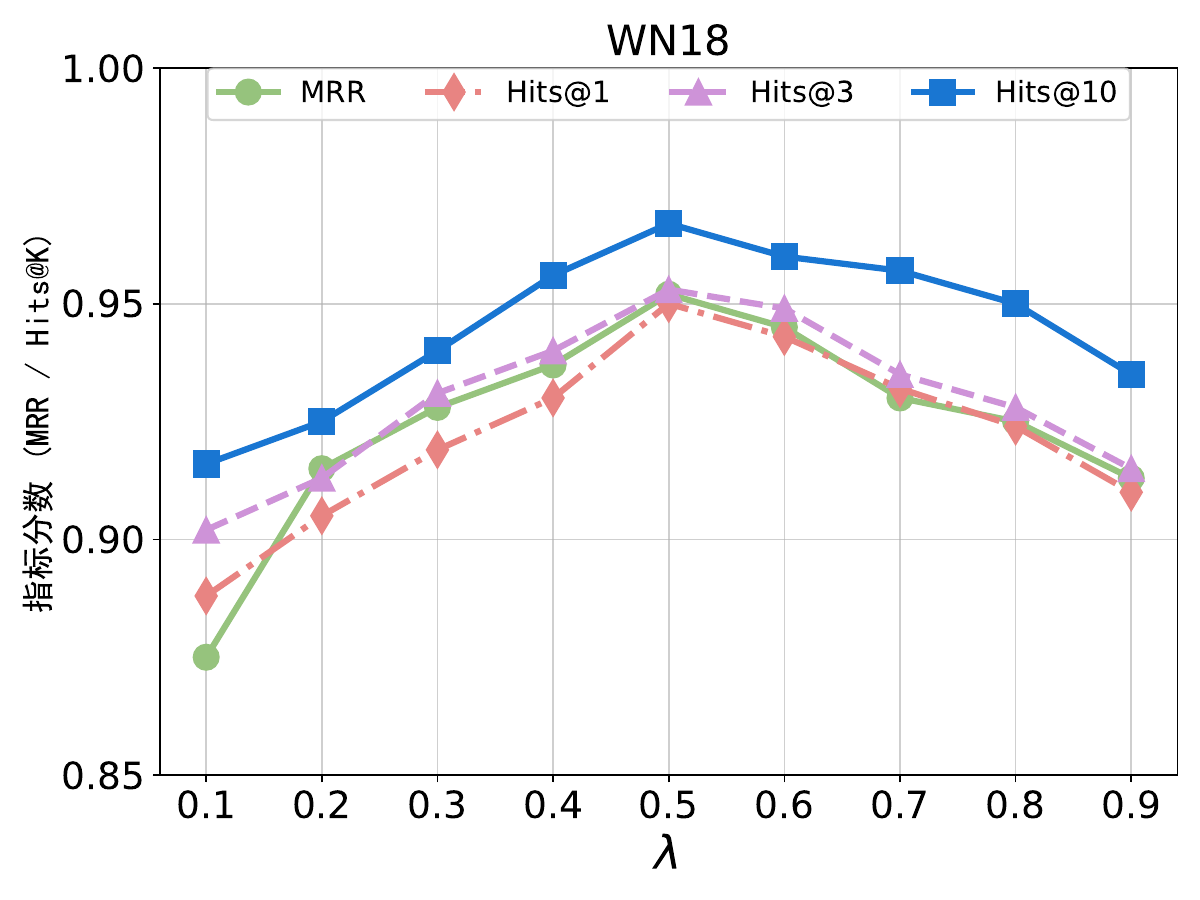} 
		}\hfill
		\subfloat[WN18RR]{%
			\label{fig:lambda_WN18RR}
			\includegraphics[width=0.48\linewidth]{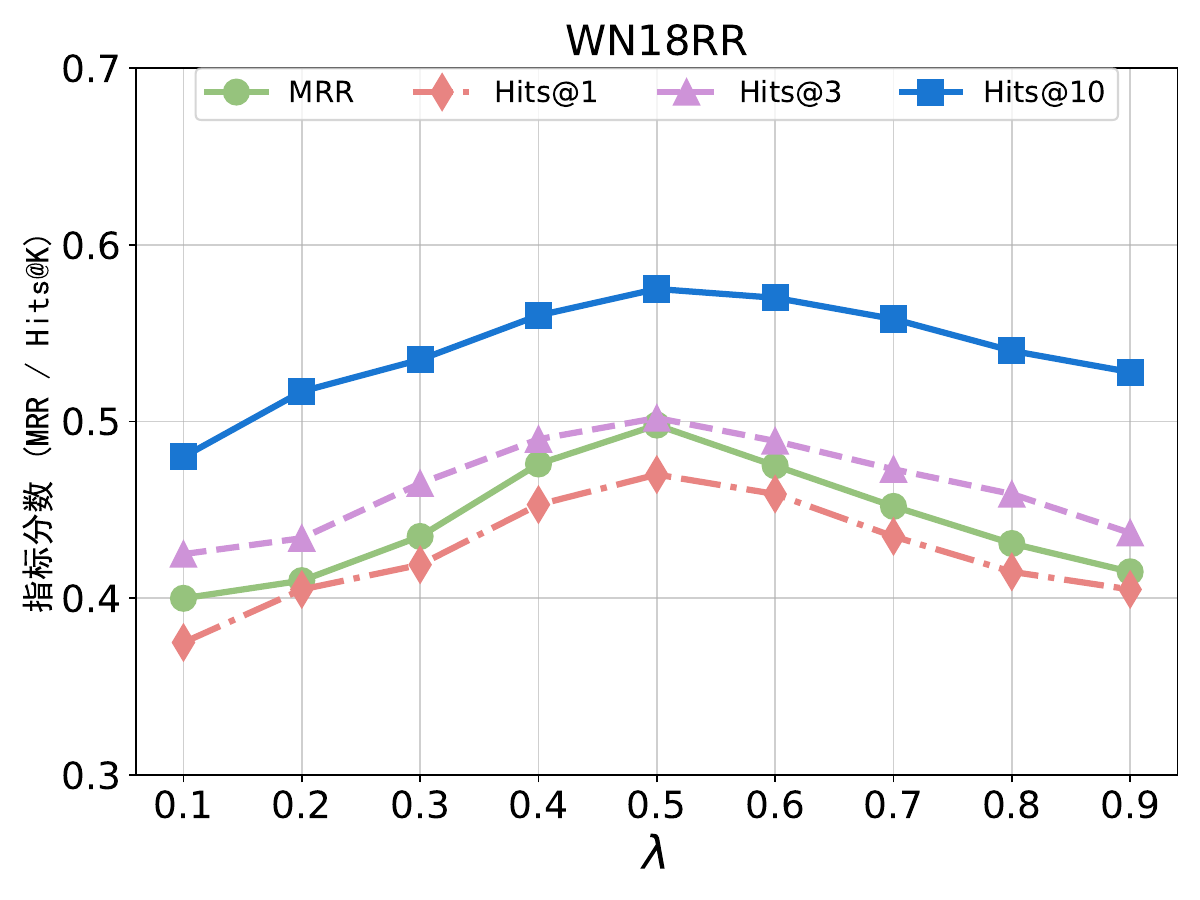} 
		}
		
		\vspace{0.5em} 
		\subfloat[FB15k]{%
			\label{fig:lambda_FB15K}
			\includegraphics[width=0.48\linewidth]{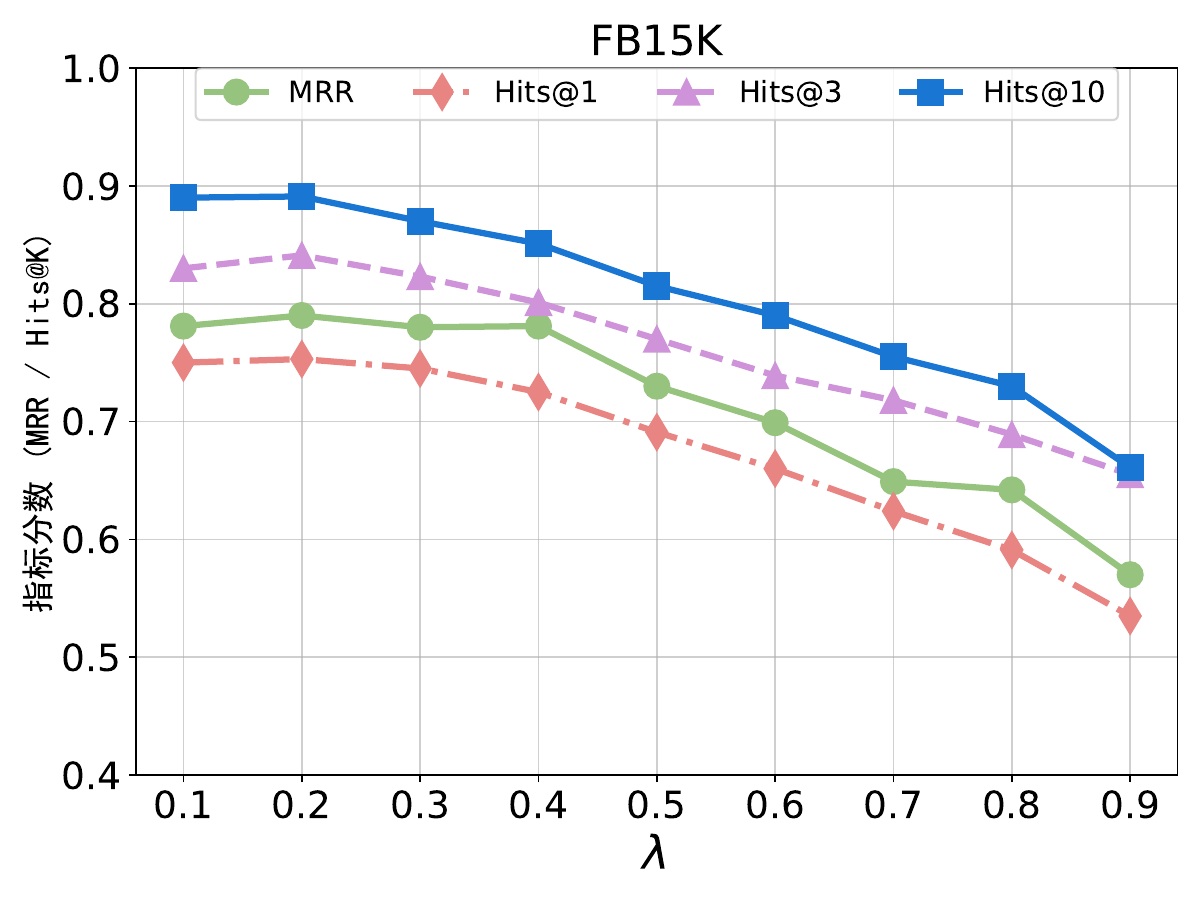} 
		}\hfill
		\subfloat[FB15k-237]{%
			\label{fig:lambda_237}
			\includegraphics[width=0.48\linewidth]{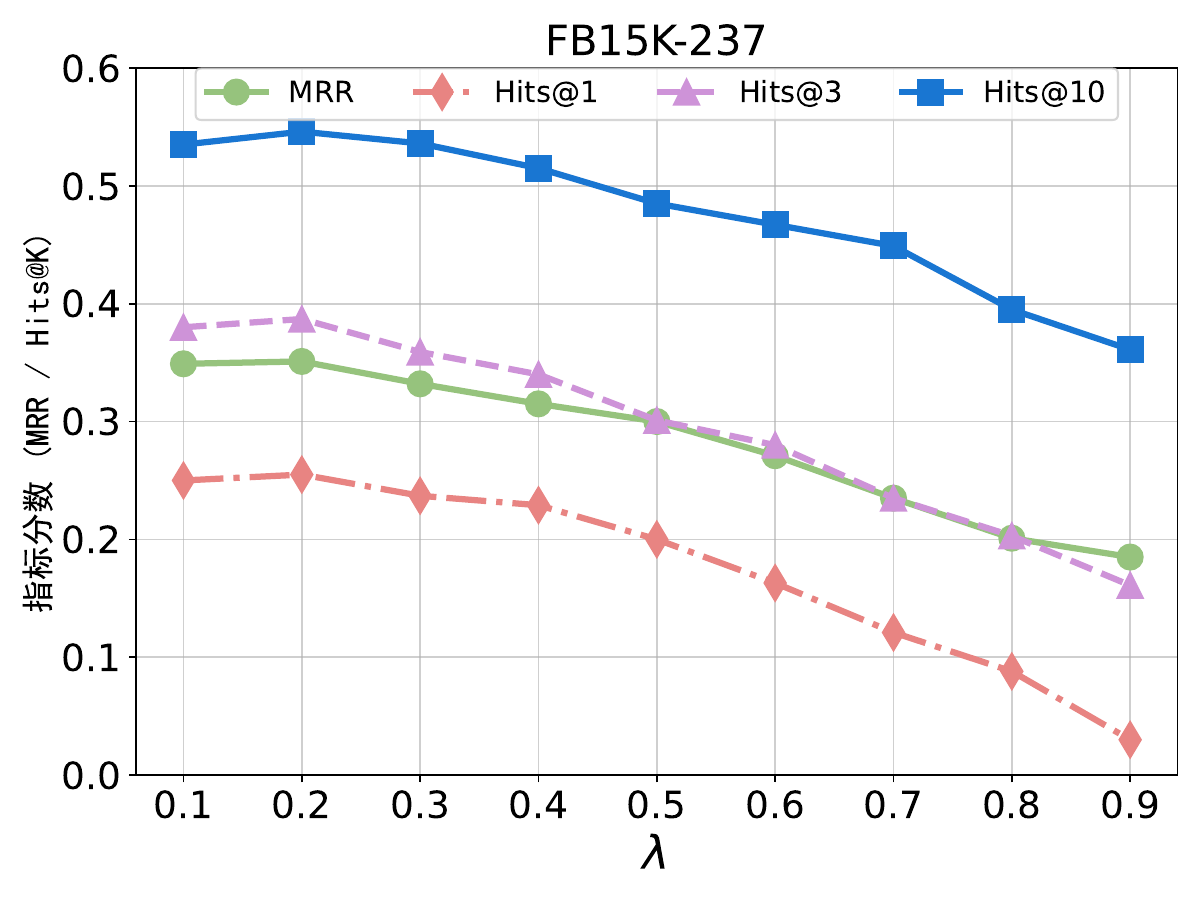} 
		}
		
		\caption{Parameter sensitivity analysis for the width regularization coefficient $\lambda$.}
		\label{fig:lambda}
	\end{figure}
	
	The experimental results evaluated via MRR and Hits@K are visualized in Figure \ref{fig:lambda}. On the WordNet datasets (sparse graphs), the performance curves exhibit a textbook inverted U-shape. As $\lambda$ increases from 0.1 to approximately 0.5, all metrics steadily climb to their global peaks. However, when $\lambda > 0.6$, performance begins to degrade. This suggests that sparse structures tolerate width regularization relatively well; they require moderate width constraints to prevent region over-inflation, but excessive penalties disrupt performance.
	
	Conversely, on the FB15k series, the model achieves its peak performance at relatively small $\lambda$ values. As the penalty intensifies, all metrics exhibit a rapid monotonic decline. This behavior originates from the dense nature of FB15k graphs, which contain massive many-to-many interactions. Under these conditions, a strong width penalty forces relation regions to shrink excessively, forcefully squeezing valid, complex entities out of the target region. Thus, CORE is highly sensitive to width penalties in dense KG topologies.
	
	\subsection{Ablation Study}
	To empirically validate the contribution of the two primary components designed in CORE, we conduct ablation studies targeting: 1) the torus space property (\textbf{t}), where its removal degrades the space back to standard Euclidean space; and 2) the dynamic context-aware entity representation mechanism (\textbf{b}) equipped with transformational bumps. The results are summarized in Table \ref{tab:ablation_all}.
	
	\begin{table*}[htbp]
		\centering
		\small 
		\setlength{\tabcolsep}{4pt} 
		\caption{Ablation Study Results on the Four Benchmark Datasets}
		\label{tab:ablation_all}
		
		\vspace{0.5em}
		\centerline{(a) WN18 and WN18RR Datasets}
		\vspace{0.5em}
		\begin{tabular*}{\linewidth}{@{\extracolsep{\fill}} cc *{8}{c} @{}}
			\toprule[1.5pt]
			\multirow{2}{*}{\textbf{t (Torus)}} & \multirow{2}{*}{\textbf{b (Bump)}} & \multicolumn{4}{c}{\textbf{WN18}} & \multicolumn{4}{c}{\textbf{WN18RR}} \\
			\cmidrule(lr){3-6} \cmidrule(lr){7-10}
			& & MRR & H@1 & H@3 & H@10 & MRR & H@1 & H@3 & H@10 \\
			\midrule[1pt]
			\checkmark &            & 0.947 & 0.940 & 0.950 & 0.954 & 0.452 & 0.422 & 0.458 & 0.512 \\
			& \checkmark & 0.925 & 0.905 & 0.912 & 0.932 & 0.451 & 0.400 & 0.472 & 0.540 \\
			\checkmark & \checkmark & \textbf{0.952} & \textbf{0.950} & \textbf{0.953} & \textbf{0.967} & \textbf{0.498} & \textbf{0.470} & \textbf{0.502} & \textbf{0.575} \\
			\bottomrule[1.5pt] 
		\end{tabular*}
		
		\vspace{1.5em} 
		
		\centerline{(b) FB15k and FB15k-237 Datasets}
		\vspace{0.5em}
		\begin{tabular*}{\linewidth}{@{\extracolsep{\fill}} cc *{8}{c} @{}}
			\toprule[1.5pt] 
			\multirow{2}{*}{\textbf{t (Torus)}} & \multirow{2}{*}{\textbf{b (Bump)}} & \multicolumn{4}{c}{\textbf{FB15k}} & \multicolumn{4}{c}{\textbf{FB15k-237}} \\
			\cmidrule(lr){3-6} \cmidrule(lr){7-10}
			& & MRR & H@1 & H@3 & H@10 & MRR & H@1 & H@3 & H@10 \\
			\midrule[1pt]
			\checkmark &            & 0.733 & 0.674 & 0.771 & 0.831 & 0.305 & 0.211 & 0.325 & 0.484 \\
			& \checkmark & 0.723 & 0.689 & 0.751 & 0.820 & 0.337 & 0.235 & 0.374 & 0.538 \\
			\checkmark & \checkmark & \textbf{0.790} & \textbf{0.753} & \textbf{0.841} & \textbf{0.891} & \textbf{0.351} & \textbf{0.255} & \textbf{0.387} & \textbf{0.546} \\
			\bottomrule[1.5pt]
		\end{tabular*}
	\end{table*}
	
The data reveals that the complete CORE architecture achieves peak performance across all evaluation metrics on every dataset. This implies that the synergy between the torus embedding space and the dynamic entity representation mechanism significantly elevates the model's link prediction capabilities.

Specifically, embedding within the torus topology exhibits the most profound impact. When the torus property is removed—degenerating the model back to an ordinary Euclidean space—performance sharply plummets across all datasets. This is most dramatically observed on FB15k, where the MRR nose-dives from 0.790 to 0.723.

Moreover, the necessity of the dynamic representation mechanism is evident when handling complex relational reasoning. When the transformational bump is disabled, performance decay is particularly stark on the highly dense FB15k-237 dataset (e.g., a 4.6\% drop in MRR). This verifies that equipping the model with relation-aware dynamic shifts grants local adaptability, seamlessly steering entities into their correct logical target regions. 

In conclusion, neither the global spatial inclusiveness of the torus manifold nor the local semantic specificity of the dynamic entity mechanism can be omitted. The integration of both design elements fundamentally drives the robust performance of CORE.

\section{Conclusion}
\label{sec:conclusion}

In this paper, we proposed CORE (Cyclic Orthotope Relation Embedding), a novel region-based embedding approach for knowledge graph completion. By embedding entities and relations into a torus manifold and innovatively representing relations as cyclic orthotopes, CORE effectively overcomes the gradient vanishing and spatial truncation issues caused by absolute boundaries in traditional Euclidean models. This boundary-less topological design guarantees smooth optimization and preserves logical expressiveness when embeddings cross spatial limits, thereby enhancing overall training efficiency and stability. 

Furthermore, we redefined the orthotope distance metric on the torus and designed an adaptive width regularization loss to prevent unconditional region inflation. Theoretical analysis strictly proves that CORE is fully expressive and capable of modeling a diverse set of complex relational patterns, including subsumption, intersection, and mutual exclusion. Extensive experiments on multiple benchmark datasets demonstrate that CORE achieves outstanding performance across various evaluation metrics, with ablation studies confirming the critical necessity of both the torus topology and the dynamic entity representation modules. 

Ultimately, the cyclic region embedding paradigm introduced in this work provides a robust and fresh spatial geometric perspective for reasoning over large-scale, complex knowledge graphs. Future work will explore the scalability of this boundary-less framework on even larger datasets containing highly diverse and intricate relational structures.


\end{document}